\documentclass{article}

\usepackage{arxiv}

\usepackage[utf8]{inputenc} 
\usepackage[T1]{fontenc}    
\usepackage{hyperref}       
\usepackage{url}            
\usepackage{booktabs}       
\usepackage{amsfonts}       
\usepackage{nicefrac}       
\usepackage{microtype}      
\usepackage{lipsum}
\usepackage[pdftex]{graphicx}
\title{Fast Concept Mapping: The Emergence of Human Abilities in Artificial Neural Networks when Learning Embodied and Self-Supervised}

\newcommand*\samethanks[1][\value{footnote}]{\footnotemark[#1]}
\author{
  Viviane Clay\thanks{Corresponding author. Email: vkakerbeck@uos.de} \\
  Institute of Cognitive Science\\
  University of Osnabr\"uck\\
  Wachsbleiche 27, 49090 Osnabrück \\
   \AND
 Peter König\thanks{Shared senior authorship. Authors contributed equally.} \\
    Institute of Cognitive Science\\
  University of Osnabr\"uck\\
  Wachsbleiche 27, 49090 Osnabrück \\
  \And
 Gordon Pipa\samethanks \\
    Institute of Cognitive Science\\
  University of Osnabr\"uck\\
  Wachsbleiche 27, 49090 Osnabrück \\
  \And
 Kai-Uwe Kühnberger\samethanks \\
    Institute of Cognitive Science\\
  University of Osnabr\"uck\\
  Wachsbleiche 27, 49090 Osnabrück \\
}

\begin{document}
\maketitle

\begin{abstract}
Most artificial neural networks used for object detection and recognition are trained in a fully supervised setup. This is not only very resource consuming as it requires large data sets of labeled examples but also very different from how humans learn. We introduce a setup in which an artificial agent first learns in a simulated world through self-supervised exploration. Following this, the representations learned through interaction with the world can be used to associate semantic concepts such as different types of doors. To do this, we use a method we call fast concept mapping which uses correlated firing patterns of neurons to define and detect semantic concepts. This association works instantaneous with very few labeled examples, similar to what we observe in humans in a phenomenon called \textit{fast mapping}. Strikingly, this method already identifies objects with as little as one labeled example which highlights the quality of the encoding learned self-supervised through embodiment using curiosity-driven exploration. It therefor presents a feasible strategy for learning concepts without much supervision and shows that through pure interaction with the world meaningful representations of an environment can be learned.
\end{abstract}


\section{Introduction}
Artificial neural networks (ANNs) by now excel at many complex tasks in fields such as visual, auditory, and natural language processing and often even outperform humans \cite{alam2019survey}. However, the types of mistakes that are observed in ANNs are often very different from the mistakes humans would make. For instance, many image processing networks have been shown to be vulnerable to adversarial attacks \cite{szegedy2013,Kurakin2016,AdvAttacksReview} which means, small perturbations in the pixel values, often not even visible to the human eye, can lead to a misclassification. Also, certain natural images with easily interpretable image content to the human eye have been shown to trick most state-of-the art image classification networks \cite{Hendrycks2019} and a general overreliance of ANNs on texture instead of object shape has been demonstrated \cite{Baker2018,brendel2018approximating}. This shows that many networks do not have a true understanding of objects like humans do but often over-fit on overall color, texture, and background cues.

There are big differences between the learning process in humans and learning in ANNs which leads to differences in behavior and the tasks that they excel at \cite{Zador2019}. By focusing on the differences in embodiment and supervision, we postulate that making these factors in the learning of ANNs more similar to human learning will lead to performance and errors more similar to what we observe in humans. All living beings interact in some way with the world. This makes it possible to learn more stable concepts about the world, relations between objects, and sensory-motor contingencies \cite{Engel2013}. ANNs are often trained with no interaction with the world as well as fully-supervised, beginning with the final task and with no gradual acquisition of knowledge. This is in stark contrast to what we believe to be the case in humans. Piaget proposes the development of a child to be split into several stages \cite{piaget1952} and especially during the first stages knowledge is largely acquired through weakly-supervised interaction with the environment \cite{Piaget1928}. Although this model is not without criticism \cite{PiagetCritic} the general idea that humans develop skills gradually over their lifespan through interaction with the world seems widely accepted.

Humans have the ability to perform fast mapping, which is a phenomenon first detailed in children by Susan Carey and Elsa Bartlett in 1978. Fast mapping describes the observation that children can learn new concepts, words, or facts after minimal exposure to them. It has been demonstrated that a single exposure to a new word can be sufficient to lead to the child remembering the word a week later \cite{FastMapping}. This means the child has the ability to make an instant association between word and meaning. This ability has also been found in other species such as dogs \cite{FastMappingDog}. Some later studies found that for fast mapping to be successful, additional memory aids and specific learning conditions are needed \cite{Gurteen2011, Horst2008}. For fast concept learning of more abstract concepts in older children and adults even more cognitive mechanisms seem to be used ranging from analogical comparisons \cite{Gentner2015}, Bayesian reasoning \cite{tenenbaum1999bayesian} to abstraction by varying prior knowledge \cite{Braithwaite2015}. In general the amount of labeled examples needed for a human to learn a new word-meaning-association is several orders of magnitudes smaller than what is needed for classical ANNs \cite{Zhang2018}. Several one-shot and few-shot learning approaches have been developed for ANNs in an attempt to classify new observations based on only one or few labeled examples \cite{chen2018a,Masi2019,Wang2020,Koch2015SiameseNN}. However, the few-shot task is usually only a generalization to new classes of a task that has previously been trained for, using a large data set of other classes. This means that these approaches still require large labeled or weakly labeled data sets of objects other than the tested ones to learn a meaningful embedding space. 

\begin{figure}
\begin{center}
\includegraphics[width=0.8\linewidth]{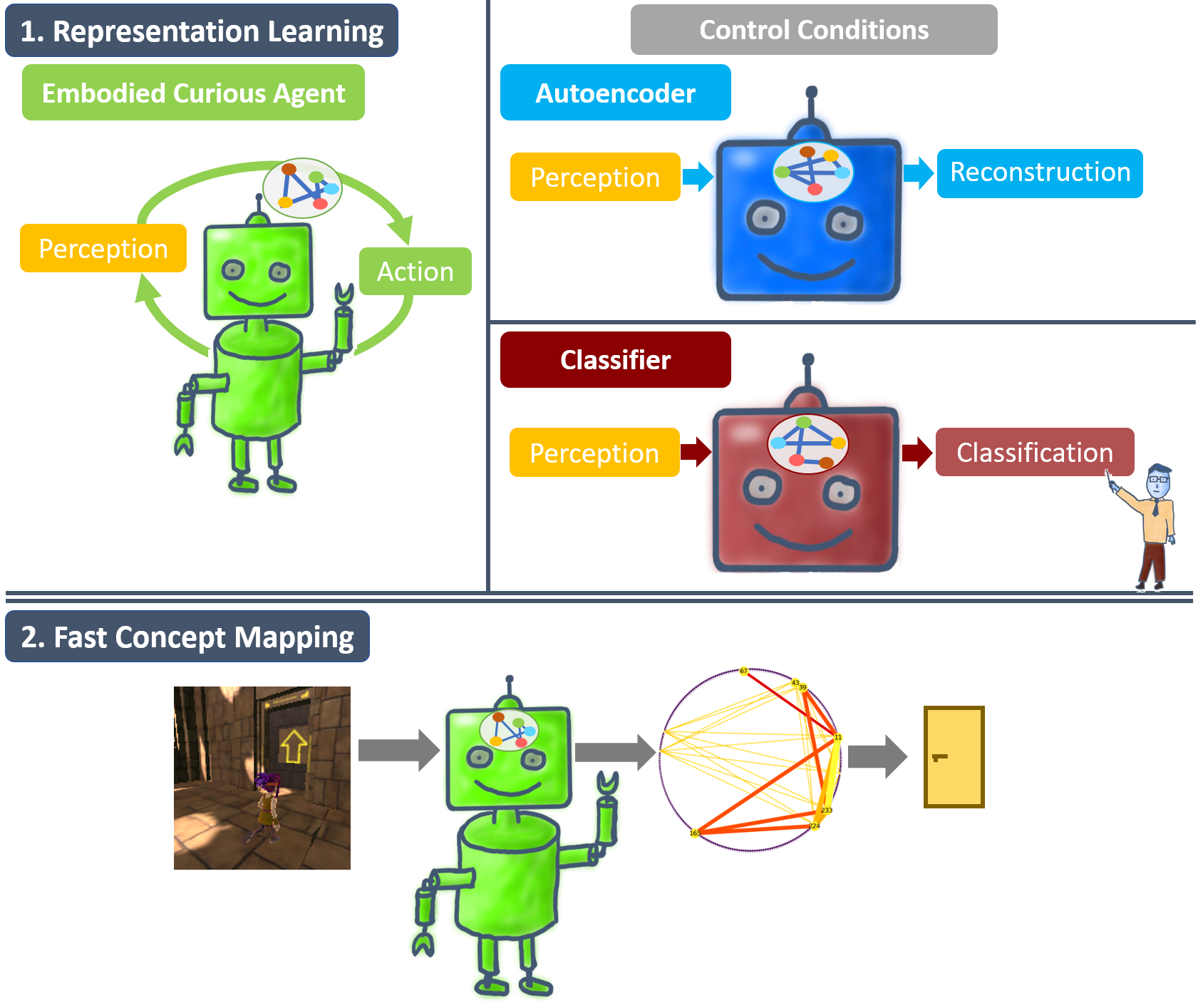}
   \caption{The learning procedure. 1. Learning an embedding of high dimensional visual input. Here we compare self-supervised learning through embodied interaction with the environment (left), a classical approach of unsupervised representation learning using an autoencoder (top right) and fully-supervised learning using an object classifier (bottom right). 2. Concept extraction from learned representations using fast concept mapping (FCM).}
   \label{learning}
\end{center}
\end{figure}

We investigate whether training an ANN more similar to how infants may learn leads to phenomena similar to those found in humans. We focus on self-supervised, embodied learning and look at the phenomenon of fast mapping. We propose to learn a meaningful embedding of visual observations purely through self-supervised interaction within a simulated world (step 1 in figure \ref{learning}, left). This interactive learning period can then be followed by a supervised, fast mapping like, association between representations and concepts using very few labeled examples (step 2 in figure \ref{learning}). Overall, this learning setup seems more similar to how we appear to learn and doesn't require large amounts of labeled training data.

\section{Methods}
\subsection{Network Structure and Training}
\label{Sec:training}
\begin{figure*}[tb]
\begin{center}
   \includegraphics[width=0.99\textwidth]{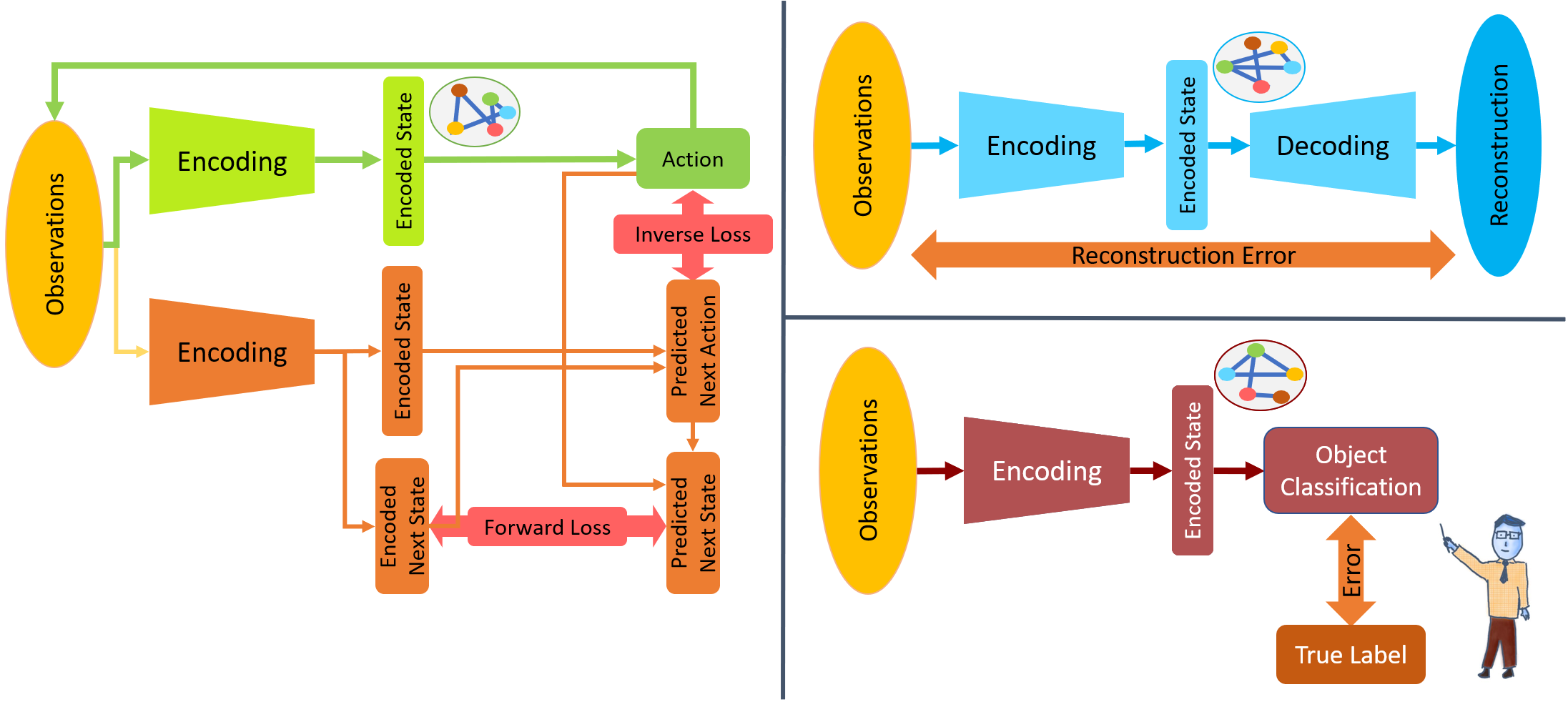}
   \caption{Network structure of the three learning conditions compared in this paper. For simplicity the individual layers are omitted. (left) Network structure of the curious agent which represents the embodied, self-supervised learning. (top right) Network structure of the autoencoder which represents non-embodied self-supervised learning. (bottom right) Network structure of the classifier which represents the fully-supervised learning condition.}
   \label{curiousStructure}
\end{center}
\end{figure*}

To demonstrate the advantages of embodied, self-supervised learning we contrast three deep neural networks with each other. All three networks have the same network architecture between input and encoded state. They differ in their output and the objective that they are trying to optimize as well as the way in which they are optimized. We take a look at two networks which are trained without any human supervision and one network trained fully-supervised. The two self-supervised conditions comprise one agent, trained using deep reinforcement learning, and one autoencoder. The agent learns through interaction with the world, outputting actions and optimizing a curiosity objective. The autoencoder learns to compress the input into an encoded state and decode this representation again, optimizing a reconstruction error between input and output. The fully-supervised network is trained on an object classification task in which the task is to output the presence of 8 different concepts in the input image which are the same concepts as above with additional option to classify image as containing 'no door' and 'puzzle piece'\footnote{No door is excluded from the analysis above as it is a negation of the other door concepts and is implicitly expressed in the door concepts not exceeding the detection threshold. Puzzle piece is also excluded due to the fact that the curious agent does not perform well enough to reach the higher levels at which the puzzles appear.}. 

All three networks receive a visual input of size 168x168x3, encode it to an encoded state of size 256 and then use this encoded state to solve their respective task. Between input and encoded state all three networks consist of two convolutional layers (16 and 32 kernels respectively) and two dense layers, each of size 256. In the following experiments we will look at the three encoded states (activations of the dense layer), how they differ, and how well we can extract semantic concepts from them.

The three networks are trained with observations collected in the obstacle tower environment \cite{Juliani2019}. The environment is a simulated 3D maze environment with a time limit and several obstacles. It consists of randomly generated levels which are made up out of several rooms connected by doors. Doors have visually marked properties such as leading to a next room, a next level, or only opening with a key or after solving a spatial puzzle. The time limit can be extended by entering new levels and by picking up blue time orbs. The visual theme can vary across levels and the illumination of different rooms is selected randomly. An agent in this environment receives a visual camera input, taken at its current position as well as a vector of size 8 with auxiliary information such as time left, current floor number, and number of keys holding. In this paper, we only look at the encoding created of the visual input. The autoencoder and the classifier receive camera images collected by a trained agent in the obstacle tower environment as their input in random order. Therefor all three networks are trained on 168x168x3 RGB images from the obstacle tower environment (examples in figure \ref{FCM} and \ref{env}) and we are taking a closer look at their respective encodings of these images after training.

The agent network shown in Figure \ref{curiousStructure} represents the embodied learning condition. When we speak of embodiment we mean, that interaction happens through a physically simulated body with a closed loop between action and perception. This loop is visualized in green here. The input observation determines which action the network produces and this action in turn influences the next observation. The actions produced by the network are discrete and divided into four action branches such that at every point in time an action in each of the branches is selected. The action branches correspond to walking (forward, backward, none), body rotation (left, right, none), camera rotation (left, right, none) and jumping (yes, no). At every time step an action is selected from each of the four action branches which means there are 54 possible action combinations. The network learns to produce actions that lead to rewards. Rewards are conventionally received from the environment for achieving certain goals such as walking through a door, entering a new level or picking up an object. Here, we want to look at learning without any external supervision so we omit all rewards from the environment and instead replace them with internal rewards. These internal rewards are produced by a curiosity module introduced in \cite{Pathak2017} and visualized in orange in Figure \ref{curiousStructure}. 

The curiosity module comprises another neural network which optimizes two objectives. The first one is to predict the next encoded state based on the current encoded state and the current action. The difference between the predicted next state and the actual next state is called the forward loss. The second objective of this network is to determine the action that was performed between the current state and the next state. The difference between the inferred action and the actual action performed is called the inverse loss. As both tasks are performed using the same encoded state, optimizing the inverse loss makes sure, that this encoded state of the curiosity network contains action relevant information. The curiosity network tries to minimize the inverse loss and the forward loss by making accurate predictions about the next state and the performed action. 

The action network (shown in green) in contrast receives the inverse loss and the forward loss as a reward weighted with 0.8 and 0.2 respectively. This enforces the action network to learn a policy which produces actions that lead to new, unpredicted observations. This implicitly rewards the network for navigating through the world, entering new rooms and new floors as these behaviors lead to new observations. An agent with a policy that stays in only one room would have a very small forward and inverse loss as the curiosity network can make very good predictions on this known environment. However, the action network would not receive many rewards from the curiosity network as it does not receive any unpredicted observations. Therefor the action networks' policy is adapted to enter through doors and explore new floors to receive the curiosity networks' rewards. Using only intrinsic curiosity the agent can learn a policy that navigates through the 3D tower environment without any external supervision.

The agent network is optimized using proximal policy optimization \cite{PPO}. Therefore, the action network does not only produce action probabilities but also a value estimate for each time step. This value estimate is used to update the network weights in a way that the cumulative reward is maximized. To keep conditions as comparable as possible and only vary the type of learning, the three networks not only have the same structure from input to encoding but also use the same optimizer (Adam optimizer \cite{kingma2017adam}) and the same batch size (256) to update their network weights. This allows us to compare between embodied, self-supervised, and fully-supervised training conditions.

\subsection{Fast Concept Mapping}

Fast concept mapping (FCM) is a simplistic approach to directly read out the encoding of a concept from a representation using a few examples of the concept. This readout only works if the concept is encoded uniquely in the representation but if it is, it can be read out with very few labeled examples.

\begin{figure}
\begin{center}
   \includegraphics[width=0.8\linewidth]{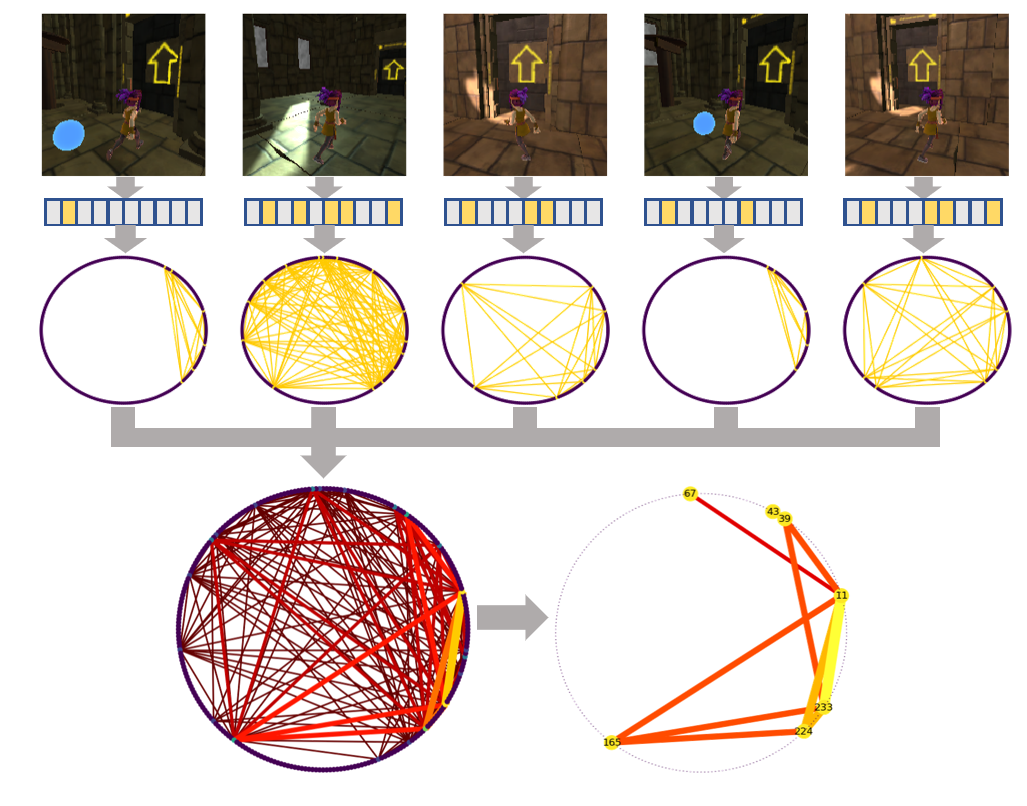}
   \caption{Fast concept mapping. Demonstrated on the concept 'level door', using five example images of the concept and the encodings of the trained agent network. First, the encodings corresponding to the input images are extracted and rewritten as a connectivity graph (top rows). Next, the sum of the five connectivity graphs is calculated (bottom left). Color represents the connection strength which is the number of examples in which the connection is present. The concept is then defined be the N strongest connections, here ten\protect\footnotemark, and their weights (bottom right).}
   \label{FCM}
\end{center}
\end{figure}

To read out a concept using FCM a number of labeled examples of the concept are needed. Figure \ref{FCM} shows FCM on the example of the concept 'level door' which is a door with a yellow arrow that leads to the next level of the environment. The example images should be different from each other and representative of the concept. It is no problem if also other concepts are present in some of the examples. In the example shown here, five instances of the concept 'level door' are used to extract the concept definition. To do this, the five examples are given to the encoding network and the encoded representation of the images is extracted. Next, for each pair of neurons that is active together in each of the five representations we assume a connection between them. If we only have one encoding, then this means that all active neurons in the encoding have connections to each other. In the following step, the sum of all connections from the example input encodings is calculated. This means, that neurons that are active together in multiple of the example inputs now have a stronger connection to each other. To define the concept 'level door' the N strongest connections are taken from the summed up connectivity graph. In this example, we take the ten strongest connections. These ten pairs of neurons as well as the normalized strength of their connection now define the concept for 'level door'. The same procedure can be repeated for any other concept one would like to extract from the encoding, resulting in a connectivity definition for each concept.

\begin{figure}
\begin{center}
   \includegraphics[width=0.8\linewidth]{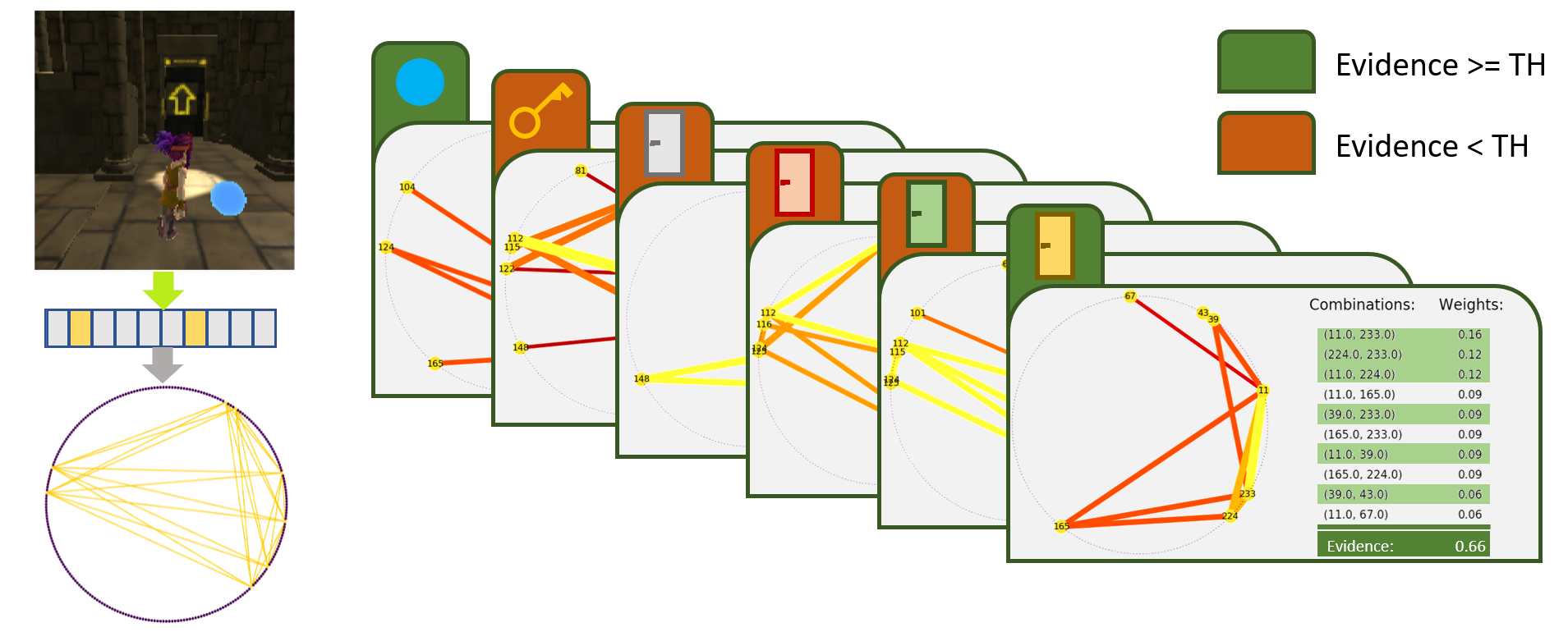}
   \caption{Fast concept mapping inference. To detect concepts in an unlabeled input the connectivity graph is extracted from the encoding corresponding to the input (left). Each concept can then be compared to this graph and every connection from the concept definition present in the new input adds more evidence for the concept. How much evidence is added is determined by the weight of the connection in the concept definition. If the evidence is above a threshold, the concept is classified as present in the new input (right).}
   \label{FCMInf}
\end{center}
\end{figure}

\footnotetext{It may be difficult to see all ten connection in the figure due to their overlaps. In Figure \ref{FCMInf} (bottom right) all ten connections are listed together with their connection strengths.}

To detect the extracted concepts in a new, unlabeled input, one first needs the connectivity graph for this new input. This graph can be obtained in the same way as during concept extraction, by getting the encoding of the input from the trained network and making a connection between every pair of neurons that is active together in response to the input. To compare the new inputs' graph with the concept definition one looks for how much evidence for the concept is found in the new graph. This is done by adding up the normalized weights of every connection in the concept definition that can also be found in the new graph. If every connection from the concept definition would also be present in the new graph, meaning that all pairs of neurons in the definition are active in response to the new input, then the evidence for this concept would be one. If no connection from the concept definition is present then the evidence for this concept is zero. A threshold is set, which determines how much evidence for a concept needs to be found in a new input encoding for this concept to be classified as present in the input. An investigation into the effect of the threshold on concept detection is provided in the next section.

In the example shown in Figure \ref{FCMInf}, we find that six of the ten connections in the concept definition for 'level door' are also active in the test image encoding shown on the left side. If we now add up the weights for these six connections we get an evidence of 0.66 for this concept in the image. As this is above our threshold, we classify the level door as present in the image. We now repeat the same procedure on all the other concepts that were extracted and detect additionally to the level door the concept 'blue time orb'. For the other concepts we do not find enough evidence in the encoding and therefor classify them as not present.

\section{Results}

\subsection{Structure in Representations Learned Through Embodiment}
All results are obtained from artificial neural networks trained on 168x168x3 dimensional visual observations from a 3D maze environment shown in the supplementary material (Figure \ref{env}) and described in the methods section\ref{Sec:training}. The investigated encodings are the activations of 256 neurons in the last hidden layer before action selection for the agent, deconvolution and reconstruction for the autoencoder, and classification in the classifier. The network structure leading up to this representation is the same for all three conditions and described in details in the methods section \ref{Sec:training}.

As shown in a previous study \cite{CLAY2020} conducted in the same environment, the embodied agent learns a sparse and meaningful encoding of its high dimensional visual input. As opposed to the previous study, this is achieved without any external rewards from the environment, using only intrinsic curiosity as a learning signal \cite{Pathak2017}. In a set of 8400 frames collected in the 3D environment there are an average of 8 neurons active in each frame \footnote{Active is defined as an absolute activation bigger than the average activation of this neuron. This definition of activation compared to the original activation strengths and a universal threshold can be seen in Figure \ref{THComp} in the supplementary material. The adaptive threshold is biologically and theoretically motivated\cite{Leugering2018} and helps with the much more dense encoding of the classifier and the autoencoder.} (min=0, max=29, var=13.12) which is 3.15\% of the 256 neurons in the visual encoding. 91.02\% of the neurons in the visual encoding are active in at least one frame of the test run. The most active neuron is active in 54.71\% of the frames. This is a very sparse encoding of the 84672 dimensional visual input with a wide variety of selective activations in the hidden layer.

\begin{figure}
\begin{center}
   \includegraphics[width=0.5\textwidth]{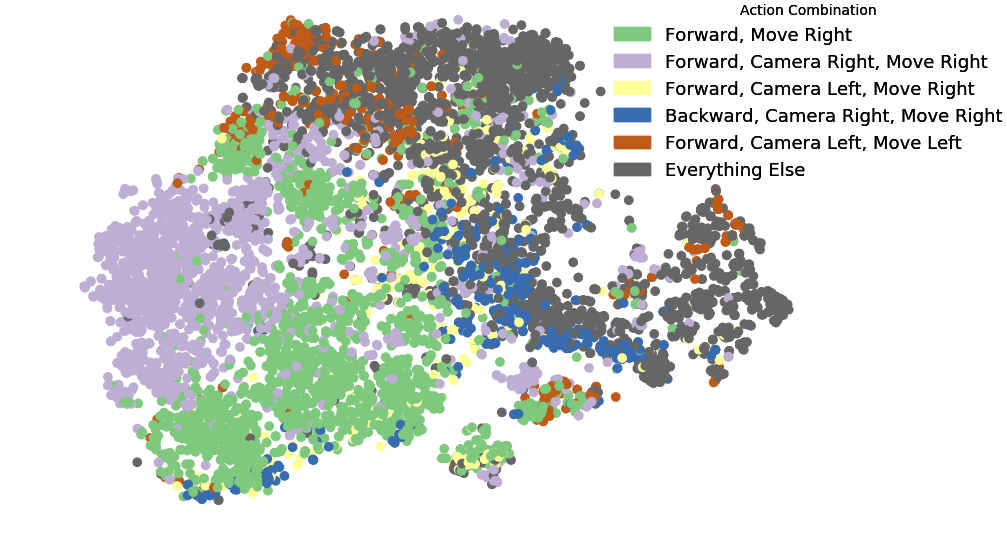}
   \includegraphics[width=0.38\textwidth]{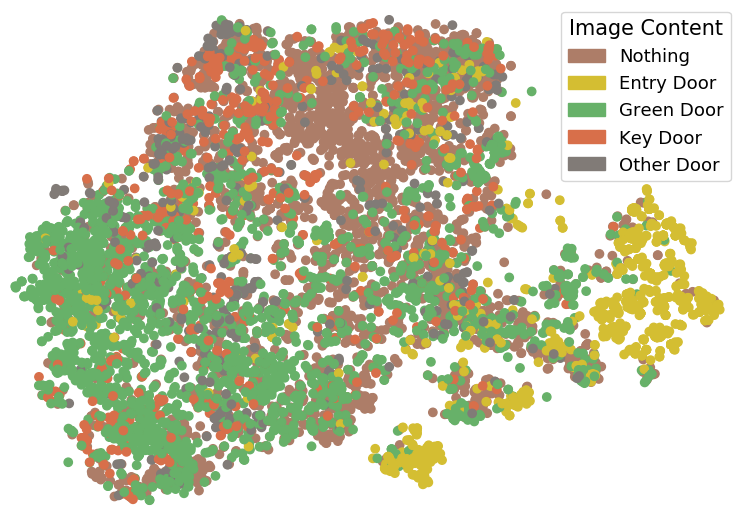}
   \caption{Meaningful structure in the image encodings of the embodied, curious agent. t-SNE projection of latent representations colored by action (left) and colored by type of door it shows (right). Each dot represents the encoding of one input image (168x168x3) in the hidden layer of the trained agent (256 neurons). The grey colored dots in action-colored projection summarize 34 action combinations. The other five colors show the five most common action combinations.}
   \label{tsne_action}
\end{center}
\end{figure}

With an unsupervised dimensionality reduction method such as t-SNE \cite{maaten2008visualizing} one can investigate whether meaningful structure can be found in the encodings. In Figure \ref{tsne_action} the 8400 test encodings of dimensionality 256 are projected into two-dimensional space using t-SNE. In such a strong reduction of dimensionality not all information can be preserved but nevertheless one can see some structure in regards to actions (left) and objects (right) in the encodings. Especially the most common action combinations as well as the level entry door are encoded distinctly even in the two dimensional projection. In our FCM approach we make use of the full dimensionality of the encodings to extract even more structure and to be able to disentangle overlap between object and action encodings.

\subsection{Fast Concept Mapping}

After demonstrating that the learned representation shows some meaningful structure in action and object space (figure \ref{tsne_action}), we want to see whether concepts can be extracted automatically from the encodings with few labeled examples. To do this we design a procedure called fast concept mapping (FCM) which is described in detail in the methods section. The general idea is to take some example images of a concept (for example five images of a door) and look at consistent activation patterns in the learned encodings of these examples. These consistent patterns then define the concept and can be detected in the activations corresponding to novel observations. Here we use five random example images for each of the six different concepts. Then we test the extracted concepts on 250 test images of the concept and 250 randomly sampled from the other concepts such that chance performance is at 50\%. We compare the performance  in detecting the extracted concepts between the embodied agent and the three control conditions. The autoencoder and the classifier representations are trained as described in methods. The random condition uses binary vectors drawn from a random uniform distribution. Performance shown in this section refers to the ability to extract concepts from the different learned representations of the visual input. It has nothing to do with the performance on the task that the networks were originally trained on. Therefor we compare here whether different training objectives (embodied and self-supervised, self-supervised, fully-supervised, none) lead to a difference in the ability to extract concepts using FCM.

\begin{figure}
\begin{center}
   \includegraphics[width=0.8\linewidth]{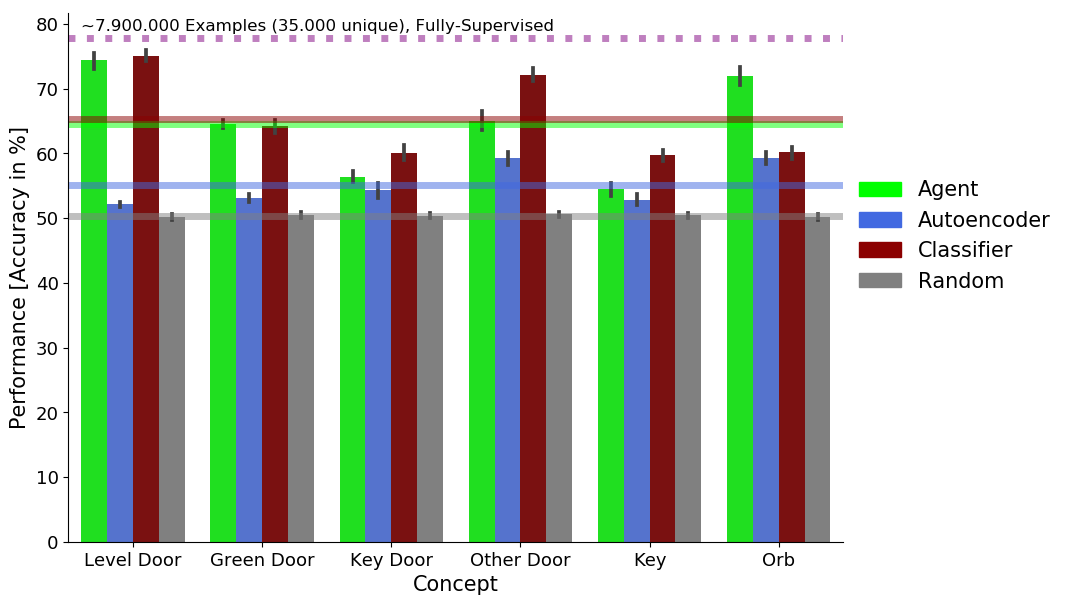}
   \caption{Accuracy for six different concepts comparing representations learned by the embodied curious agent, a classifier, an autoencoder, and a random representation. Five random example images are used to extract each concept with a pattern complexity of 10 and a threshold of 20\%. 95\% confidence intervals for 100 random sets of five examples are indicated for each bar. The dotted line shows the average performance of the classifier using its fully-supervised trained read-out layer. Horizontal lines indicate average performance over all concepts for the four conditions.}
   \label{FCMFinalStats}
\end{center}
\end{figure}

\begin{figure}
\begin{center}
   \includegraphics[width=0.8\linewidth]{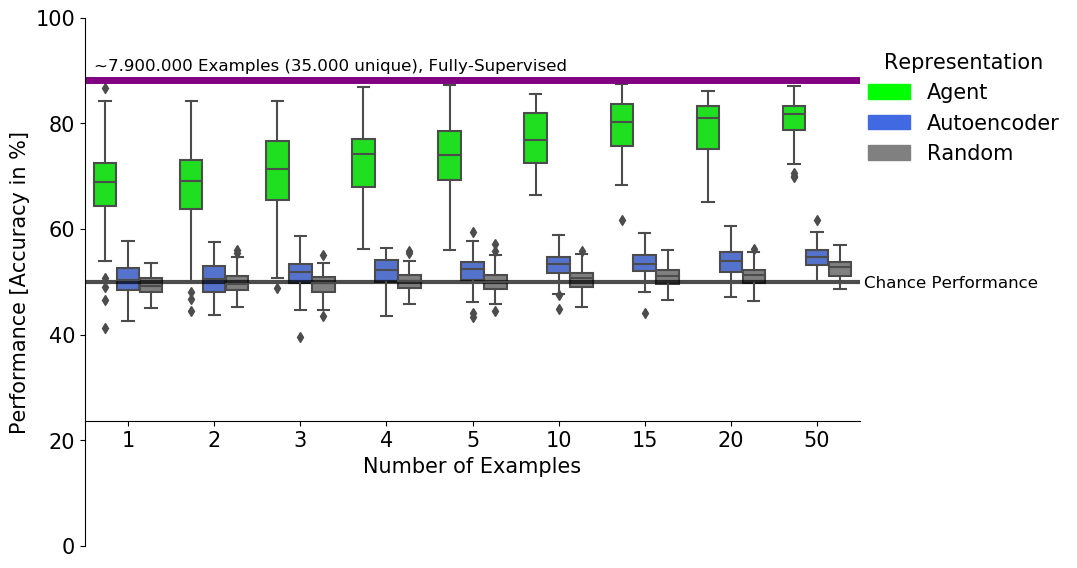}
   \caption{Performance of detecting a door (level door) in a 168x168 RGB image with 1-50 labeled examples. A comparison is shown between different learning setups used to learn the image embedding. The representation learned self-supervised through interaction with a virtual world is shown in green. A self-supervised representation learned without interaction using an autoencoder is shown in blue. Grey boxes show the baseline performance of using a random representation. The purple line indicates an upper bound performance, obtained from a fully supervised classifier trained on six million labeled examples. The boxplots show median performance as well as first and third quartile performance over 100 random sets of examples.}
   \label{numExp}
\end{center}
\end{figure}

Figure \ref{FCMFinalStats} shows the performance of FCM on the six concepts for the four representations learned under different conditions. FCM on the agent representation outperforms FCM on the autoencoder and the random representation on all six concepts. The agent's performance differences between concepts can be explained by its learning progress. Concepts that cannot be extracted well, such as the key and the key door, are not as well learned in the policy as the other four concepts. While the agent can already pick up time orbs and navigate through level doors and green doors to reach level five, it has trouble when needing to pick up a key to go through the key door, which is a task introduced at this point. The key and key door concepts do not seem to be encoded very well yet and are therefor also harder to extract. Overall, FCM works best on the agent representation and the classifier representation with some performance differences between concepts.

The comparison to the performance of the classifier read out layer, indicated by the dotted line, shows that the learned representations most likely contain more object information which could be read out with a more sophisticated method. The classifier uses a fully-supervised trained read-out layer for this, which is fitted to extract the object information using over 7.9 million labeled examples. As the output of the classifier is based on the embedding of the classifier we can see directly from the comparison between the red line and the dotted line how much improvement in performance a better read out layer gives and how much depends on the encoding.
To investigate how few examples are needed to extract the concepts from the learned representations we measure the performance when using 1, 2, 3, 4, 5, 10, 15, 20, and 50 examples. The effect of the number of labeled examples on the performance is shown in Figure \ref{numExp} for the concept 'level door'. One can see that already with only one labeled example the agent achieves a performance significantly above chance. Showing more labeled examples leads to an increase in performance for the agent. Also the autoencoder performance increases slightly with more labeled examples. However, the agent strongly outperforms the autoencoder, no matter how many labeled examples are shown. The fully-supervised classifier read-out layer outperforms the agent but has been trained with several orders of magnitude more labeled examples.

\subsubsection{Parameter Selection}

\begin{figure*}[tb]
\begin{center}
   \includegraphics[width=0.9\textwidth]{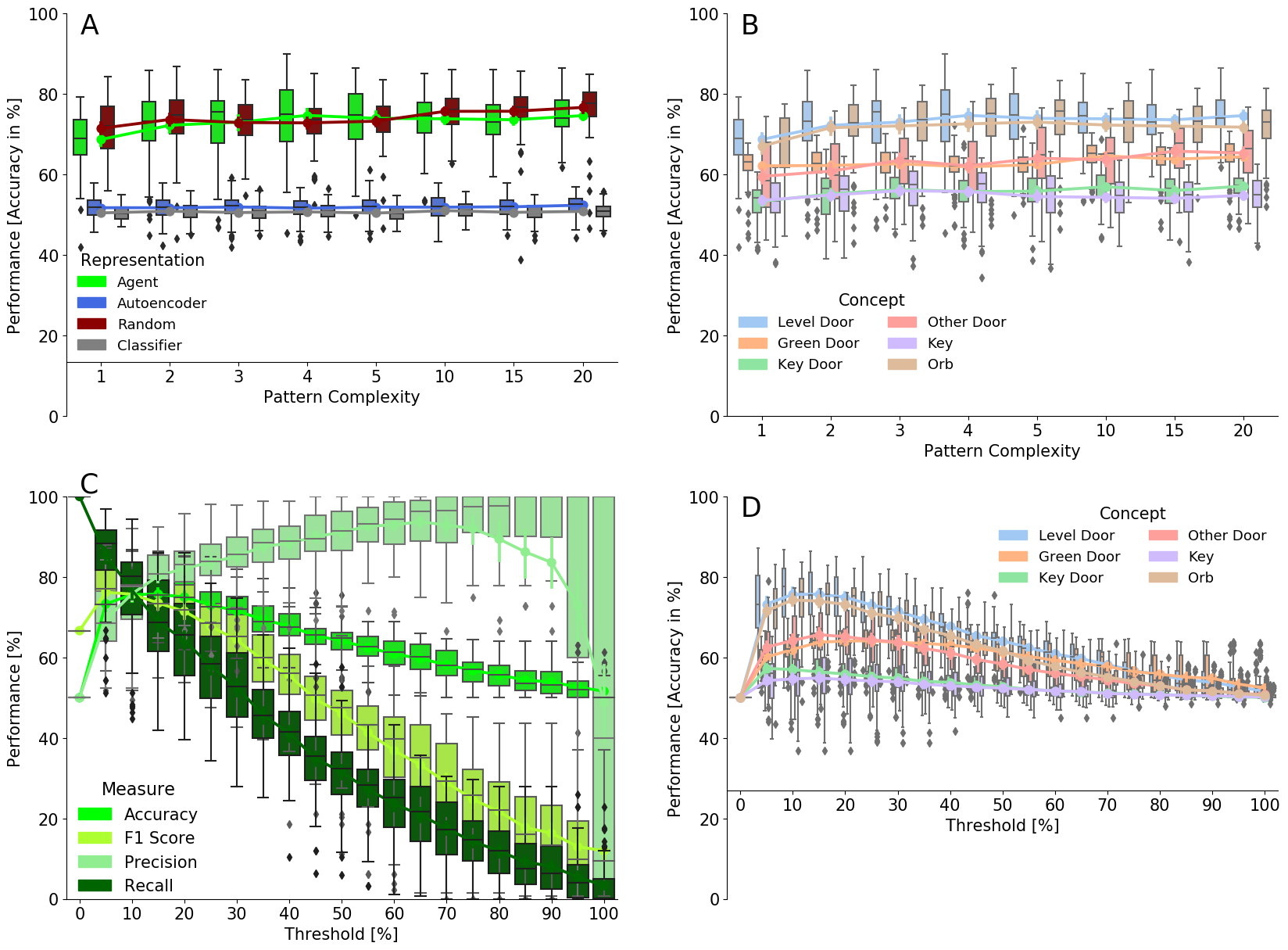}
   \caption{The effect of pattern complexity and threshold on concept mapping performance. The box plots show median performance as well as first and third quartile performance over 100 random sets of examples. The point plots show the point estimate mean and its 95\% confidence interval. For all experiments five examples are used for concept extraction. In the pattern comparison a threshold of 20\% is used. In the threshold comparison a pattern complexity of 10 is used. (A) Comparison between different representations dependent on the pattern complexity used to define the concept 'level door'. (B) Comparison of pattern complexity for each concept on the agent representation. (C) Threshold (amount of evidence needed for concept detection) comparison on the concept level door for the agent. Comparisons for the other conditions can be found in the supplementary material. (D) Threshold comparison for all concepts in the agent.}
   \label{ParaComp}
\end{center}
\end{figure*}

To perform FCM two parameters need to be selected. The first parameter is the pattern complexity which is the number of neuron pairs that is used to define the concept during concept extraction. The second one is the threshold which determines how much evidence for a concept is needed to classify it as present during inference. Finding the best values for these parameters requires more than the five examples used for the previous results. However, once the ideal parameters are found for one concept they can be used for all other concepts as we can observe the same trend in all concepts (see Figure \ref{ParaComp}, B and D). Additionally, FCM is surprisingly robust in regards to the pattern complexity parameter. As shown in Figure \ref{ParaComp} A, the choice of pattern complexity has no strong effect on the performance. Once at least two neuron pairs are used to define the concept, there is no significant effect of using a higher pattern complexity. This could be due to the encoding that the agent uses. If on average only 8 neurons are active in each frame, then it does not help much to add more neuron pairs to define a concept. 

The performance of FCM can be more sensitive to the selection of the threshold parameter. As shown in Figure \ref{ParaComp} on the bottom left, the choice of the threshold is a trade off between precision and recall. Is the threshold set low, which means that little evidence of a concept is needed to detect it, the recall is high and the precision is low. With a high threshold that requires a lot of evidence for a concept to be recognized, the recall is low and the precision is high. When considering the F1 score as a good balance between precision and recall, the best threshold is between 10\% and 20\%. When comparing the ideal threshold for the different concepts (Figure \ref{ParaComp}, bottom right) a similar trend for all concepts can be observed. However, there are some concepts where the threshold has little effect on performance (for example key door and key). These concepts therefor could not be extracted effectively from the representation.

\section{Discussion}
The most obvious advantage of the method introduced here is the small amount of labeled examples that is required to learn a concept, with an above chance performance with as little as one labeled example. As opposed to other few-shot learning methods \cite{chen2018a,Masi2019,Wang2020,Koch2015SiameseNN}, one does not only require very few examples of the test classes but also the representation learning works without any labeled examples. This makes the learning comparable to what we observe in humans. An embodied, curiosity driven learning period leads to a meaningful representation of the world, which can then be used to perform new tasks such as object detection with few supervised learning examples.

The fast concept mapping method introduced here is a fast association between consistent neuron firing patterns and semantic concepts. It works instantaneously without the need for weight optimization. The method of looking for correlated firing patterns that persist over different instances of the same concept is simple enough that it could be implemented with biological neurons\cite{Magee2020,Leugering2018}. The success of this method is based on the assumption that the representation that it is applied to uses a unique and consistent encoding of the concept. We showed that a self-supervised, embodied agent can learn such an encoding of action relevant concepts. Extracting a new concept does not require any retraining and does not have an effect on the concepts that have already been extracted. 

Even though in theory, arbitrarily many concepts can be extracted if they are encoded, this is not so easy in practice. The embodied learning phase can require long training times and due to the absence of supervision it cannot be explicitly chosen which and how many concepts will be encoded. Learning in this unsupervised embodied setup is incredibly difficult and sample efficient learning in more complex environments is still an area of active research \cite{Botvinick2019}. However, human learning also takes place over a very long time frame, which may lead to more stable and robust representations that avoid current pitfalls of artificial neural networks.

A natural next step would be to use the extracted concepts and introduce them back into the embodied learning setup as an additional aid in the decision making process. This can on the one hand, help with learning a more efficient policy due to the compressed information of concepts in the observations. Additionally it could lead to more refined encodings of the concepts themselves and a higher accuracy in detecting them. Ultimately, it could achieve language grounding without explicit enforcement through the constant presence of a language grounding task, as it is currently done \cite{hermann2017grounded, hill2019understanding, hill2020grounded}.

Another extension could be to perform reasoning on the levels of the concepts. Some concepts, such as the different types of doors, are subconcepts of a higher level concept. Other concepts can be combined with each other, such as 'left' and 'right' or 'far away' and 'close' can be combined with the different objects. Some concepts can be applied to certain types of objects, for example doors can be 'opened' or 'closed'. Due to the independent populations of active neurons in response to different concepts in a good encoding, all these concepts could be extracted independently using the FCM method and put into logical relations, either automatically deduced from activity overlap or by hand.

\section{Conclusion}
Overall, we have shown that representations learned through embodiment with no external supervision encode meaningful information about the content of high dimensional visual input. These representations can be associated with semantic labels about equally well as representations that were optimized fully-supervised for object detection. This association works fast and robustly with few randomly chosen labeled examples, similar to the ability of children to perform fast mapping. We find that concepts such as different types of doors and other action-relevant objects are encoded even though no semantic concepts were ever explicitly taught during training. Our results show the viability of a new approach to train ANNs, inspired by the way humans seem to learn. This approach focuses on self-supervised learning through embodied interaction and only transitioning later on to supervised concept learning with few labeled examples.

\section*{Acknowledgments}
The project was financed by the funds of a research training group provided by the DFG (ID GRK2340).

\section*{Author Contributions}
V.C. conceived the idea for the experiments and the fast concept mapping procedure, implemented the training, FCM and analysis and took the lead in writing the manuscript. All authors discussed the experiment design and results and contributed to the final manuscript. P.K., G.P and K.K. supervised the project.

\section*{Competing Interests Statement}
The authors declare no competing interests.

\bibliographystyle{unsrt}  
\bibliography{references}  

\newpage
\appendix
\section{Supplementary Material}
\subsection{Environment}
\begin{figure}[h]
\begin{center}
   \includegraphics[width=0.8\linewidth]{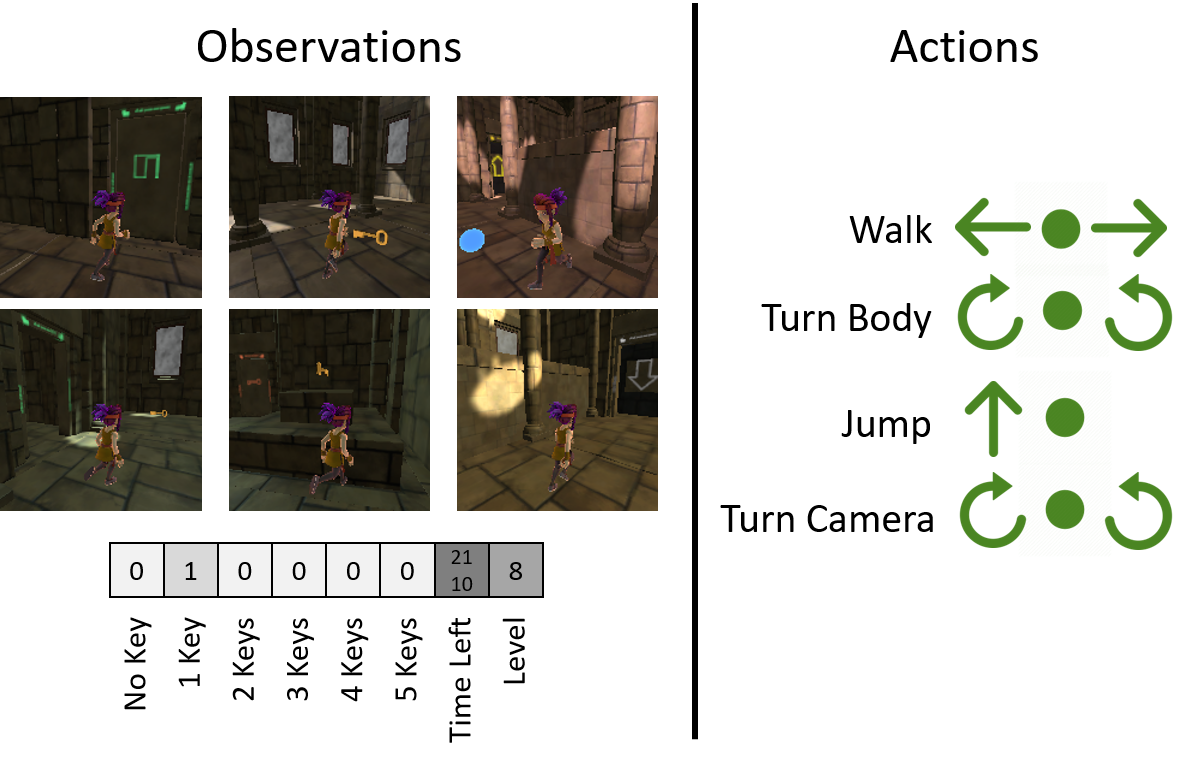}
   \caption{The Obstacle Tower environment. (left) Six examples for the visual observations. In reading direction: a green door, a key, a level door and an orb, a key and a green door, a key and a key door, an other door. Below is the structure of the vector observations which is a eight dimensional vector containing information about the number of keys holding, the time left and the current level. (right) All possible actions. The actions are divided into four branches (rows here) and at each time step one action is selected from each of the branches. Branch 0, 1 and 3 have three options each. The second branch has two options (jump or do not jump).}
   \label{env}
\end{center}
\end{figure}

\subsection{Training Specifics}

\begin{table}[h]
\begin{center}
\begin{tabular}{|l|l|}
\hline
PPO Parameter & Value \\ \hline
Batch Size & 256 \\ \hline
Beta & 5.0e-3 \\ \hline
Buffer Size & 1024 \\ \hline
Epsilon & 0.2 \\ \hline
Gamma & 0.999 \\ \hline
Hidden Units & 256 \\ \hline
Lambda & 0.9 \\ \hline
Learning Rate & 1.0e-4 \\ \hline
Normalize & True \\ \hline
\# Layer & 2 \\ \hline
Use Recurrency & False \\ \hline
\end{tabular}
\end{center}
\caption{PPO training parameters for the embodied agent (in all three reward conditions). Code for agent training can be found here: \url{https://github.com/vkakerbeck/ml-agents-dev}. Code for the autoencoder and classifier training as well as the analysis can be found here: \url{https://github.com/vkakerbeck/FastConceptMapping}. Trained models and the labeled test set can be found in \cite{Clay2020Data}: \url{http://dx.doi.org/10.17632/zdh4d5ws2z.2}.}
\label{PPOParameter}

\end{table}

\newpage
\subsection{Definition for Neuron Activation}

\begin{figure}[h]
\begin{center}
   \includegraphics[width=\linewidth]{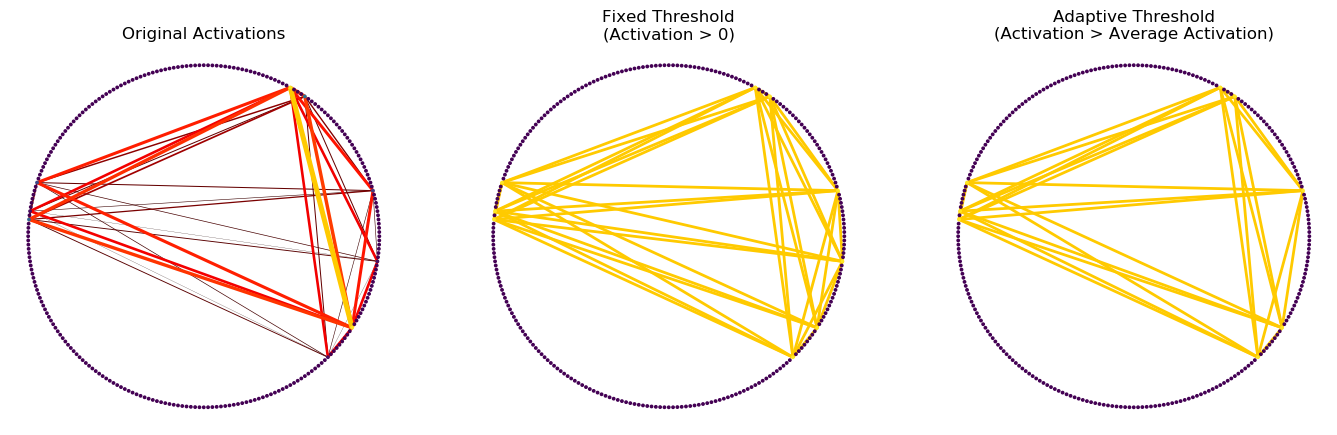}
   \includegraphics[width=\linewidth]{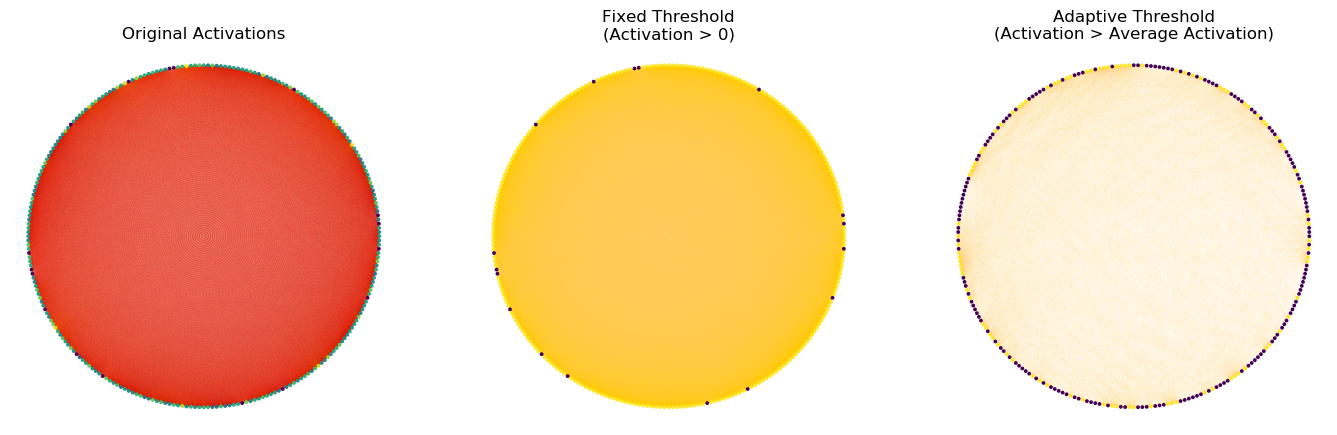}
   \includegraphics[width=\linewidth]{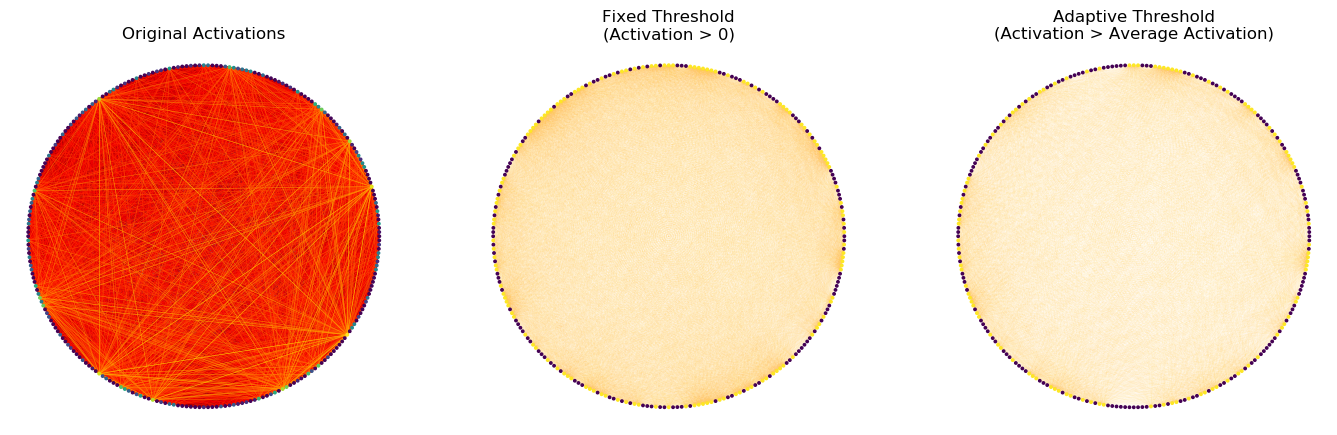}
   \caption{Original Activations (left), and thresholded actiations with a fixed threshold (middle) and an adaptive threshold (right) for one example image in the agent (top), the autoencoder (middle) and the classifier (bottom). For the FCM results shown here an adaptive threshold is used.}
   \label{patternType}
\end{center}
\end{figure}

\newpage
\subsection{Threshold Selection}
\begin{figure}[h]
\begin{center}
   \includegraphics[width=0.9\linewidth]{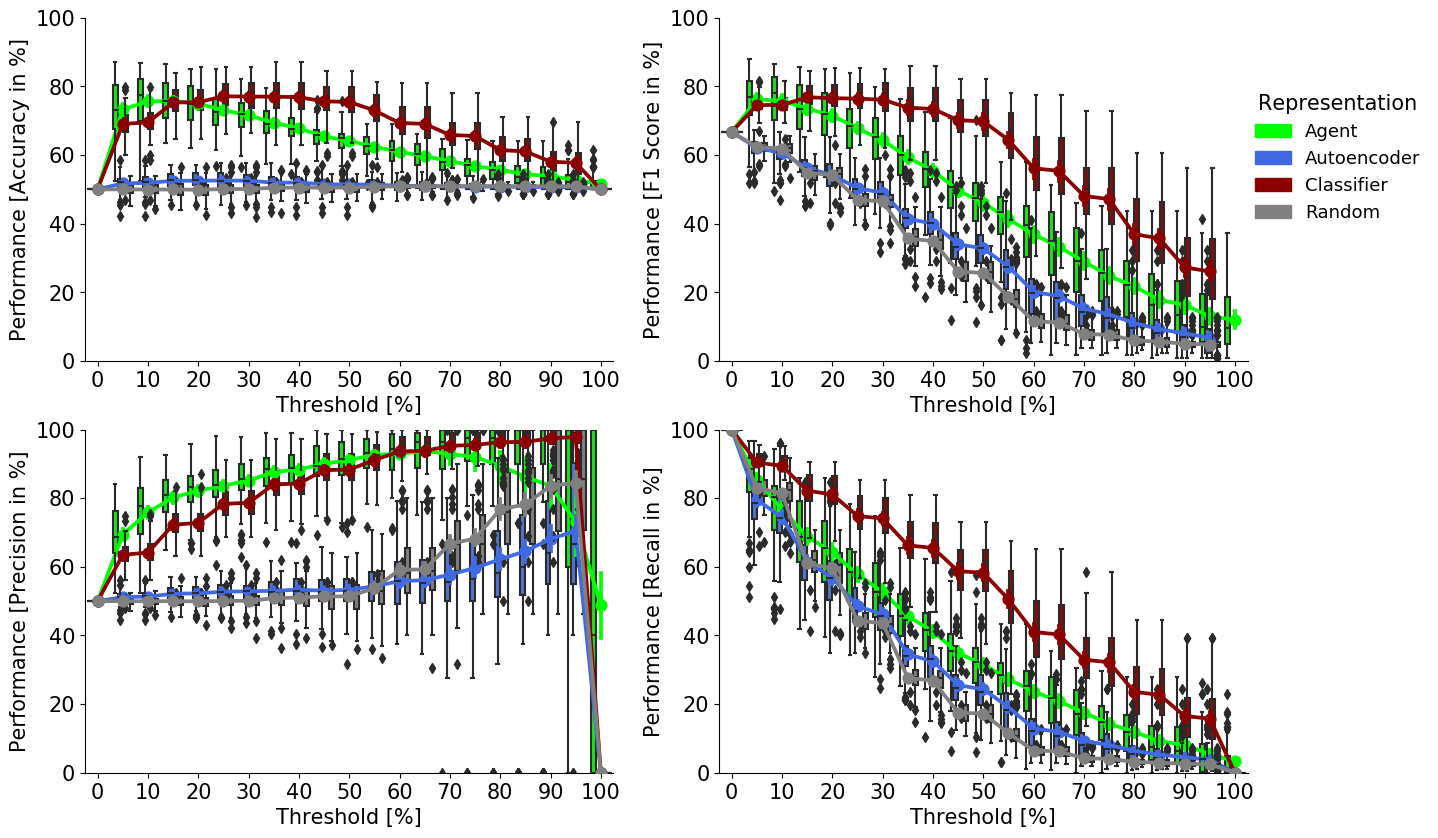}
   \caption{Threshold comparison for all representations. Accuracy, F1 score, precision and recall for different thresholds for the concept level door. The point plots show the point estimate mean and its 95\% confidence interval.}
   \label{THComp}
\end{center}
\end{figure}

\newpage
\subsection{Type of Pattern}
\begin{figure}[h]
\begin{center}
   \includegraphics[width=0.7\linewidth]{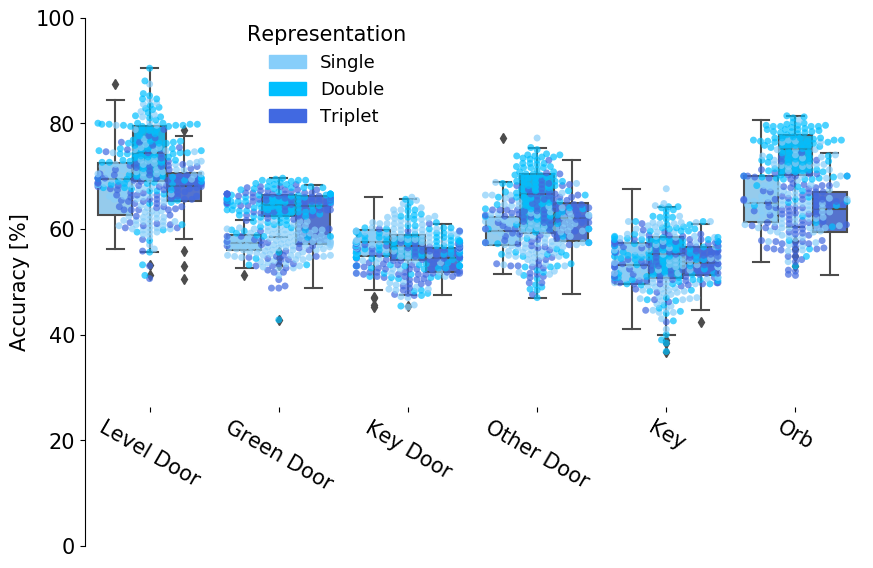}
   \caption{Comparison between different types of patterns (single activations, doubles, and triplets). The box plots show median performance as well as first and third quartile performance over 100 random sets of examples. The box plots show median performance as well as first and third quartile performance over 100 random sets of examples. }
   \label{patternType}
\end{center}
\end{figure}

\subsection{Comparison to Support Vector Machine}
\begin{figure}[h]
\begin{center}
   \includegraphics[width=0.7\linewidth]{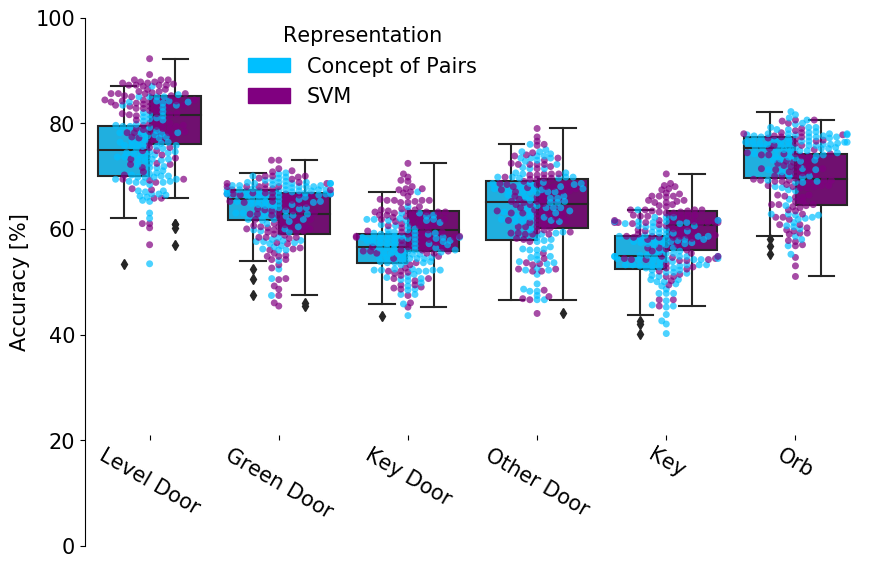}
   \caption{Comparison between using pair activations for concept definition (as described in the main text) and using a support vector machine (SVM). The SVM uses five positive and five negative examples as opposed to the FCM method which does not require negative examples and therefor only uses 5 positive examples of each concept. The box plots show median performance as well as first and third quartile performance over 100 random sets of examples.}
   \label{SVM}
\end{center}
\end{figure}

\newpage
\subsection{Supervision}
\begin{figure}[h]
\begin{center}
   \includegraphics[width=0.7\linewidth]{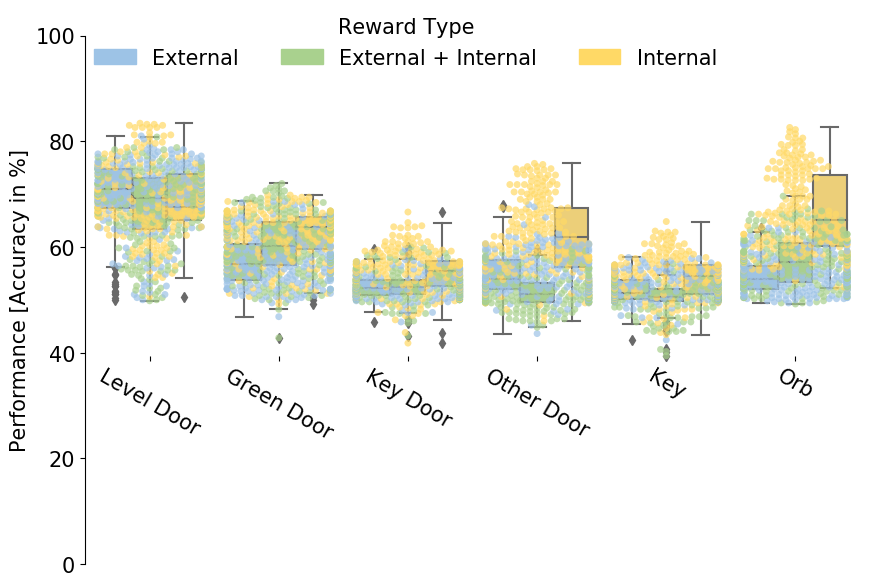}
   \caption{Comparison between three agents trained with different degrees of supervision. External supervision receives sparse rewards from the environment (1 for entering a new level and picking up a key, 0.1 for picking up a time orb and passing through a normal door). Internal supervision only received reward from the agents curiosity module as described in the main paper. External + internal receives environment rewards and curiosity rewards. The box plots show median performance as well as first and third quartile performance over 100 random sets of examples.}
   \label{SVM}
\end{center}
\end{figure}

\end{document}